\documentclass{article}

\usepackage{microtype}
\usepackage{graphicx}
\usepackage{subfig}
\usepackage{booktabs} 

\usepackage{hyperref}

\usepackage{enumitem}
\setenumerate[1]{itemsep=0pt,partopsep=0pt,parsep=\parskip,topsep=5pt}
\setitemize[1]{itemsep=0pt,partopsep=0pt,parsep=\parskip,topsep=5pt}
\setdescription{itemsep=0pt,partopsep=0pt,parsep=\parskip,topsep=5pt}

\usepackage[accepted]{icml2024}

\usepackage{amsmath}
\usepackage{amssymb}
\usepackage{mathtools}
\usepackage{amsthm}

\usepackage{makecell}
\usepackage{bbding}
\usepackage{xspace}
\usepackage{multirow} 
\usepackage{multicol} 
\usepackage[capitalize,noabbrev]{cleveref}

\theoremstyle{plain}

\theoremstyle{definition}

\theoremstyle{remark}

\usepackage[textsize=tiny]{todonotes}

\icmltitlerunning{Submission and Formatting Instructions for ICML 2024}

\begin{document}

\twocolumn[
\icmltitle{Language-assisted Vision Model Debugger: A Sample-Free Approach to Finding and Fixing Bugs}



\icmlsetsymbol{equal}{*}

\begin{icmlauthorlist}
\icmlauthor{Chaoquan Jiang}{yyy}
\icmlauthor{Jinqiang Wang}{yyy}
\icmlauthor{Rui Hu}{yyy}
\icmlauthor{Jitao Sang}{yyy}
\end{icmlauthorlist}

\icmlaffiliation{yyy}{
School of  Computer  and  Information 
Technology \& Beijing Key Lab  of 
Traffic Data Analysis and  Mining, 
Beijing  Jiaotong University,  Beijing, China}

\icmlcorrespondingauthor{Jitao Sang}{jtsang@bjtu.edu.cn}

\icmlkeywords{Machine Learning, ICML}

\vskip 0.3in
]



\printAffiliationsAndNotice{}  

\begin{abstract}
Vision models with high overall accuracy often exhibit systematic errors on some important subsets of data, posing potential serious safety concerns. Diagnosing such bugs of vision models is gaining increased attention, but traditional diagnostic methods require labor-intensive data collection and attributes annotation. Also, a recent study has demonstrated the cross-modal transferability of the vision-language embedding space and it shows the visual classiﬁers on top of the space can be diagnosed through natural language. Motivate by these developments, we propose the debugger LaVMD, which enables the vision-language model to proxy the buggy vision model by aligning the embedding spaces without any labeled visual data. Our proposed debugger can discover underperforming subgroups (or \emph{slices}) in vision models and further improve performances of the subgroups \textbf{through natural language only}. To avoid prior knowledge needed to select candidate visual attribute set, we use a Large Language Model (LLM) to obtain attributes set and probe texts, as the proxy for images to probe visual bugs. Finally, we demonstrate the effectiveness to diagnose underperforming subgroups of buggy vision models using only language on the Waterbirds , CelebA and NICO++. We can identify comprehensible attributes to users and improve buggy model by texts data.
\end{abstract}

\section{Introduction}
\label{sec:intro}
Deep vision models have shown good performance in a wide variety of fields, and been applied in many high-stakes scenes like medical image \cite{OakdenRayner2019HiddenSC,Kermany2018IdentifyingMD}, autonomous driving \cite{Cui2018MultimodalTP}, and facial recognition \cite{Seo2021UnsupervisedLO}. However, these models are also known to make numerous errors on important subgroups of data arising from shortcuts and spurious correlation \cite{Geirhos2020ShortcutLI}. These failed subgroups share certain task-irrelevant similar attributes or concepts (e.g. backgrounds) differing from shortcuts. For example, a vision model can correctly detect cows on grass background, while cows are not detected when they appear in an uncommon scene (e.g. on the beach) \cite{Beery_2018_ECCV}. Moreover, such errors maybe cause safety problems and broad societal impact. Therefore, model diagnosis, aimed to gain a systematic and comprehensive understanding of when and why models fail \cite{zhang2023drml}, has become increasingly crucial. 
\begin{figure}[t]
  \centering
\includegraphics[width=1.0\linewidth]{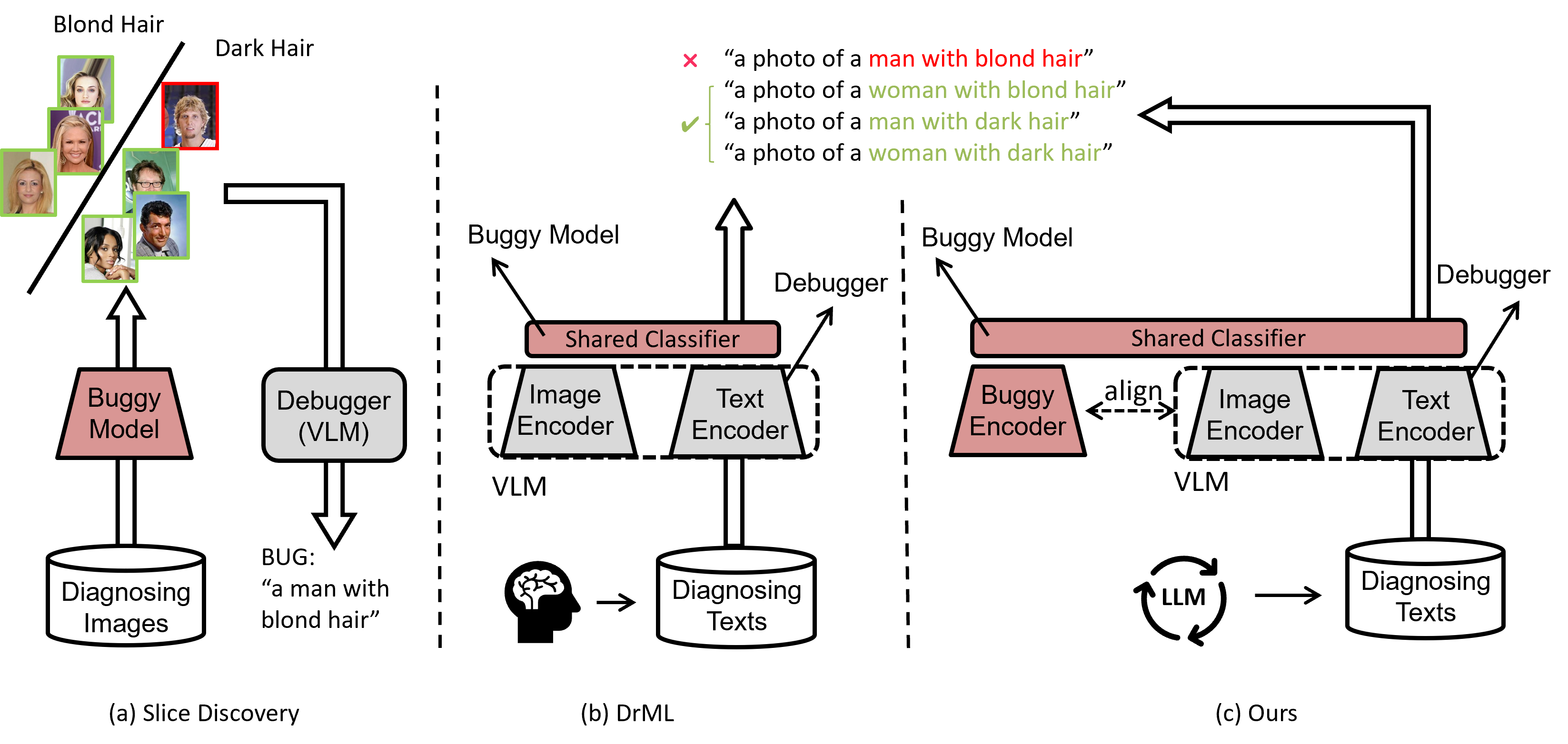}
   \caption{Comparison of different diagnostic solutions. (a) automatically clustering and describing visual failures using image data, (b) using language to prob visual failures on top of a multimodal embedding space, and (c) our method, using language to prob visual failures of arbitrary buggy model.}
   \label{fig:headshot}
\vskip -0.2in
\end{figure}
 In the practice of machine learning fairness, one could collect and manually annotate image sets along predefined bias dimensions to bias-auditing \cite{Holstein2018ImprovingFI,karkkainenfairface}. For example, the CelebA dataset \cite{Liu2014DeepLF} for face recognition has rich attribute annotations (e.g. male, age) accompanying each example to test fairness. However, traditional data collection methodologies are expensive to perform and and limit the scope of diagnosis to predefined candidate biases. 
 
Many recent diagnostic approches are designed to address the above challenge -- \emph{lack of annotated visual data}, which requires extensive data collection and human annotations. Some prior works annotate spurious features via limited human supervision \cite{Singla2021SalientIH,Singla2020UnderstandingFO}, but these are challenging to large scale datasets. Also, the discovered spurious features may not correspond to a human-identiﬁable visual attribute \cite{2021BMVC_UDIS}. Recent works pose the problem as automatically discovering and describing failure subgroups by vision-language models (called \emph{slice discovery}), as illustrated in \cref{fig:headshot}(a) \cite{Chung2018SliceFA,dEon2021TheSA,Eyuboglu2022DominoDS,Jain2022DistillingMF,Kim2023BiastoTextDU}. For instance, Eyuboglu et~al. cluster and interpret the failure subgroups as language via the vision-language model \cite{Radford2021LearningTV}, Jain et~al. \cite{Jain2022DistillingMF} distill failure directions using SVM in the embedding space, or Kim et~al. extract and annotate attributes of failure examples using image captions \cite{Kim2023BiastoTextDU}. However, these methods rely on visual examples of failure modes (e.g. the subgroup of the man with blond hair in hair-color classification task), which potentially requires large-enough labeled datasets to capture failure modes, as failure modes may be rare concepts or long-tail data distributions. 

In this work, our goal is to automatically diagnose failures of vision models and avoid collecting labeled data. Inspired by DrML \cite{zhang2023drml}, which has demonstrated the cross-modal transferability of multi-modal contrastive representation space, we suggest to discover bugs in the model using language instead of images. Our proposed method first aligns the embedding space of the buggy vision model by the CLIP \cite{Radford2021LearningTV} to obtain a cross-modal debugger; then the debugger can diagnose failures using language by sharing last layer linear classifier, as illustrated in \cref{fig:headshot}(c). Although the DrML \cite{zhang2023drml} has also used language to test performance of vision classifiers onc top of vision-language representation space \cite{Radford2021LearningTV}, as illustrated in \cref{fig:headshot}(b), its ability is limited to diagnose other vision models and it requires manually predefined attribute set to generate test texts. We addressed these challenges by allowing CLIP to proxy other vision models to accept text inputs, and querying a Large Language Model (LLM) to obtain candidate attributes. Our proposed debugger has three practical advantages: \textbf{arbitrary model}, \textbf{no labeled images}, and \textbf{no candidate attributes} -- e.g. diagnosing any pretrained image classification model provided by a third party, requiring language and unlabeled images, automatically obtaining common visual concepts or attributes related to the vision task without prior knowledge. The comparison with automatically discovering and describing visual failures (e.g. FD \cite{Jain2022DistillingMF}, B2T \cite{Kim2023BiastoTextDU} and Domino \cite{Eyuboglu2022DominoDS}) and earlier cross-modal diagnosis DrML \cite{zhang2023drml} is illustrated in \cref{tab:head_comparision}. Our debugger can be used in more practical scenarios.

\begin{table}
  \caption{Advantages comparison between our proposed practical diagnostic method and other automated methods.FD \cite{Jain2022DistillingMF}, B2T \cite{Kim2023BiastoTextDU} and Domino \cite{Eyuboglu2022DominoDS} are automatical slice discovery methods, DrML \cite{zhang2023drml} is a cross-madal diagnosis method.
  }
\label{tab:head_comparision}
\vskip 0.15in
    \centering
  \begin{tabular}{@{}lccc@{}}
    \toprule
    Method & \makecell[c]{Arbitrary\\model}&\makecell[c]{No labeled\\ images}& \makecell[c]{No candidate\\attributes}\\
    \midrule
    FD & \CheckmarkBold & \XSolidBrush & \CheckmarkBold\\
    B2T & \CheckmarkBold & \XSolidBrush & \CheckmarkBold \\
    Domino& \CheckmarkBold & \XSolidBrush & \CheckmarkBold \\
    \midrule
    DrML &\XSolidBrush & \CheckmarkBold & \XSolidBrush \\
    Ours & \CheckmarkBold & \CheckmarkBold  & \CheckmarkBold\\
    \bottomrule
  \end{tabular}
  \vskip -0.2in
\end{table}
Our contributions can be summarized as follows:
\begin{itemize}
\item We show that feature distillation allows Vision-Language Model (VLM) to proxy the trained vision model, enabling to recognize texts similar to the vision task;
\item We implement LaVMD (Language-assisted Vision Model Debugger), an approach to diagnose vision models directly using language only, without any labeled image data and candidate visual attributes;
\item We  illustrate  the  practical  application of our approach by showcasing how it interprets model decisions as language and repairs buggy models.
\end{itemize}

\section{Related Works}
\label{sec:related_work}
\textbf{Slice Discovery:}
Deep vision models often make systematic errors on specific subsets of data.
Slice discovery methods aim to identify biased groups of data that share similar attributes and exhibit consistent (often, underperforming) predictions. Many solutions to the problem focused on tabular data or metadata \cite{Chung2018SliceFA,Sagadeeva2021SliceLineFL}. In the absence of structured data annotations, some autonomous methods cluster the failed samples \cite{dEon2021TheSA} in the embedding space and describe them using natural languages \cite{Eyuboglu2022DominoDS,Yenamandra_2023_FACTS}. Eyuboglu et~al \cite{Eyuboglu2022DominoDS} employs Gaussian-mixture models to cluster the underperforming data slices, and Sriram et~al \cite{Yenamandra_2023_FACTS} suggest first amplifying correlations and then discoverring underperforming data slices. In addintion, Jain et~al \cite{Jain2022DistillingMF} distill failure direction using SVM in the latent spaces, while Kim et~al \cite{Kim2023BiastoTextDU} extract bias-related words from image captions. Chen et~al \cite{chen2023hibug} obtain attribute words using ChatGPT, and Zhang et~al \cite{zhang2023drml} demonstrate that an image linear classifier trained on the top of CLIP embeddings can be diagnosed by text input. These methods,which extract failure modes from failed samples, often require numerous task-related labeled images \cite{Kim2023BiastoTextDU,Jain2022DistillingMF,Eyuboglu2022DominoDS,Yenamandra_2023_FACTS,chen2023hibug}, and some do not diagnose bugs in the external pretrained model \cite{zhang2023drml}. Nevertheless, our proposed approach offers a more practical solution for diagnosis, utilizing CLIP as a diagnostic tool.

\textbf{Spurious Correlation:}
Deep learning models often exhibit robustness issues when confronted with distribution shifts, as spurious correlations may lead to significant performance degradation on certain subgroups within the data \cite{zhang2022contrastive,liu2021just,sagawa2019distributionally,idrissi2022simple}. There is a growing literature on eliminating spurious correlations, providing various approaches that relay on group labeling for per sample \cite{sagawa2019distributionally} or unsupervised machine learning algorithms \cite{seo2022unsupervised} to mitigate biases towards minority groups. Our work is closely related to identifying minority groups in the dataset, such as clustering and assigning group label for samples in the embedding space \cite{sohoni2020nosubclass} or choosing failure-related samples through a two-stage process \cite{liu2021just}. Similarly, our diagnostic approach can identify biased groups and rectify biased models.

\textbf{Multi-modal Contrastive Learning and Cross-Modality Transfer:} 
Exploring the interaction between vision and language is a fundamental research topic in artificial intelligence. Contrastive learning multimodal models, such as CLIP\cite{Radford2021LearningTV}, capture rich visual and textual features and encode them into semantically aligned embedding spaces. These image or text encoders exhibit excellent performance not only in various single-modal visual or text tasks but also in multi-modal applications like image-text matching \cite{Radford2021LearningTV, yuan2021florence}. Recent research has also extended its focus to the cross-modal transferability of the representation space. Some studies fine-tune vision models using textual data instead of images (Text as Image \cite{guo2023texts_as_image}, Text-only \cite{ nukrai2022text_only, gu2022cannt_believe})—Training textual models, such as text classification or text generation, and successfully transferring them to visual tasks like image classification \cite{dunlap2022using,guo2023texts_as_image} or image caption\cite{nukrai2022text_only}. Drawing inspiration from the transferability of the representation space of Contrastive Vision-Language Models (e.g., CLIP \cite{Radford2021LearningTV,zhang2023drml}), this work leverages CLIP \cite{Radford2021LearningTV} as a diagnostic tool to probe and diagnose buggy vision models through the lens of language.

\section{Language-assisted Vision Model Debugger}
\label{sec:method}
\begin{figure*}
\vskip 0.2in
    \subfloat[Generate candidate attributes.]{
        \label{fig:framework-a}
        \includegraphics[width=0.45\linewidth]{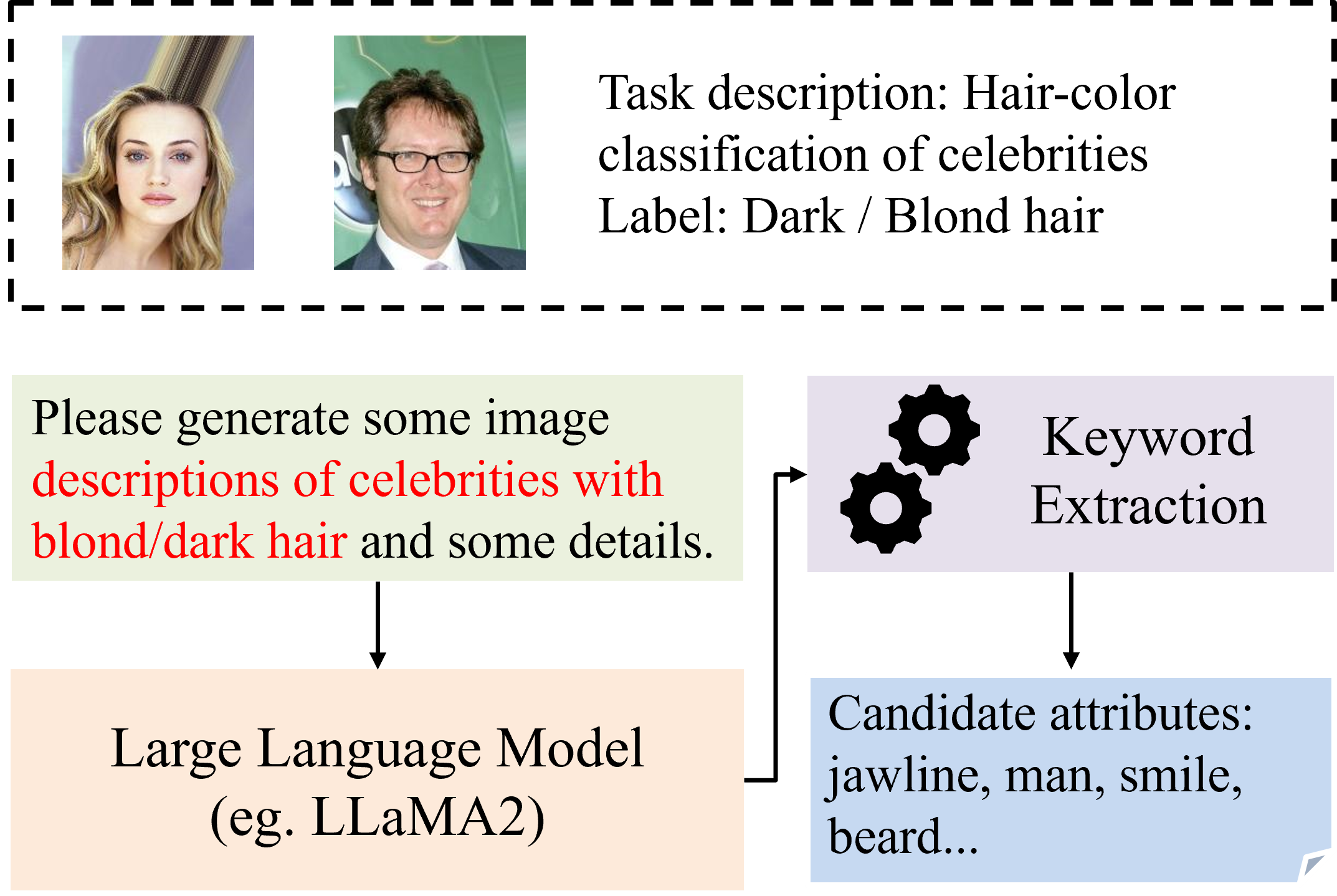}
        }\hfill
    \subfloat[Top-align with CLIP; Bottom- probe bugs using texts only.]{
        \label{fig:framework-b}
        \includegraphics[width=0.45\linewidth]{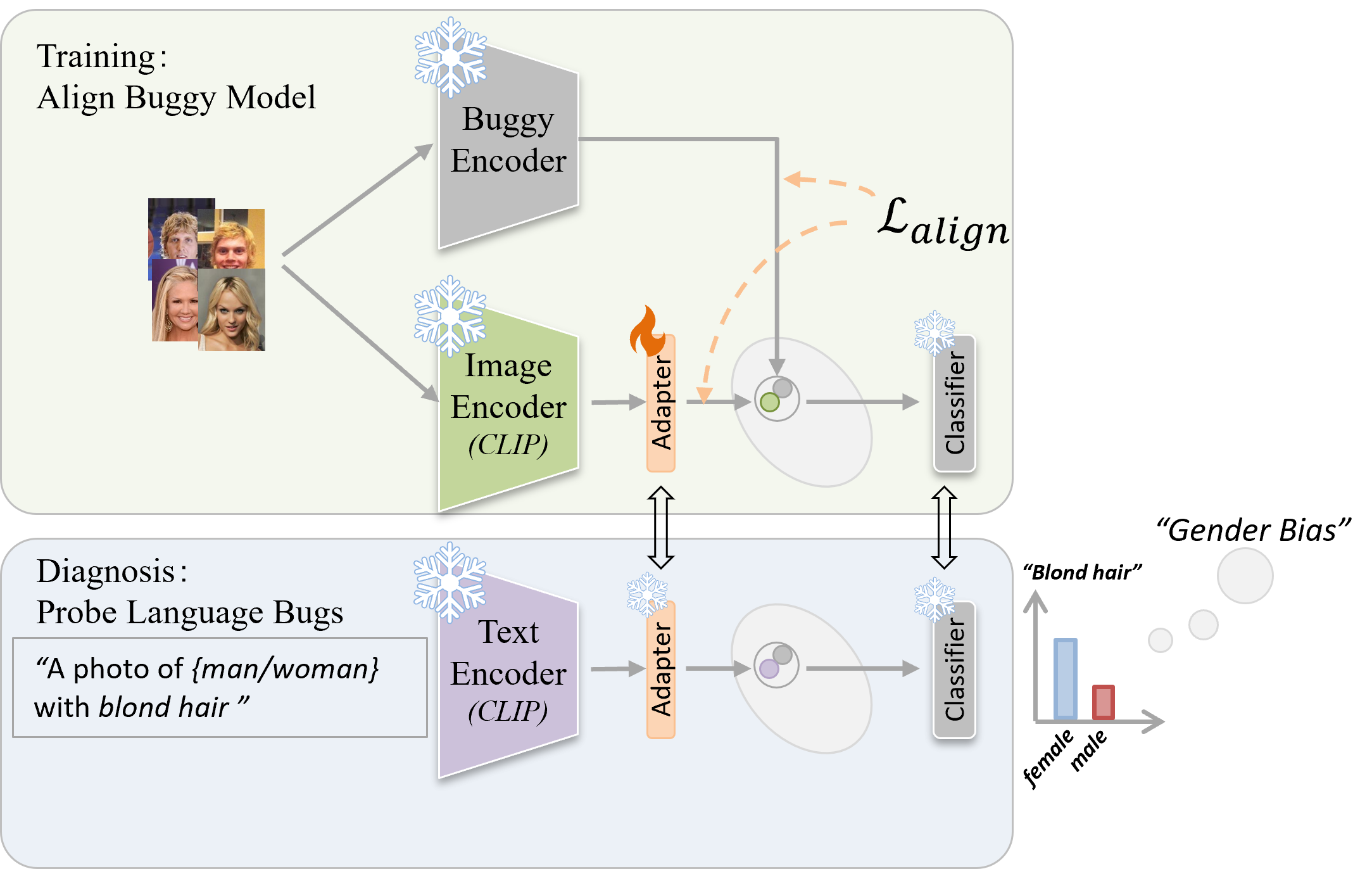}
        }
  \caption{\textbf{The framework of LaVMD}. We aim to identify underperforming subgroups of data that exhibit significantly lower performance on a task-irrelevant attribute within a category. In the example above, this refers to the bias subgroups of \emph{man} and \emph{woman} with blond hair, where \emph{man with blond hair} is a specific bias subgroup.}
  \label{fig:framework}
  \vskip -0.15in
\end{figure*}

\subsection{Symbols and Definitions}
\textbf{Discovery Error Slice:} Deep learning model often make systematic errors on subgroups of data that share some coherence attribute \cite{zhang2023drml,Eyuboglu2022DominoDS}, these subgroups are formally defined as follows:
\begin{equation}
  S = \{s\subset D \mid e(s)\gg e(D)\}
  \label{eq:error_slice}
\end{equation}
Where $e$ is the error rate, $D$ is the data set. To simplify workflow, we redefine the subgroups $s$ as combinations with labels and attributes.

Consider a standard image classification problem with input $x\in \mathcal{X}$ and label $y\in \mathcal{Y}$. In addition, each data point $(x,y)$ has some attribute $a\in \mathcal{A}=[0,1]^{k}$ that is spurious correlation with the label $y$,
such as gender and age. $\mathcal{A}$ is a characteristic or concept set and is familiar to domain expert. We can partition data into $m=|\mathcal{A} |\times |\mathcal{Y}|$ groups, each group $s=(y,a)\in \mathcal{A}\times \mathcal{Y}$. Given a pre-trained vision model $h:\mathcal{X}\rightarrow \mathcal{Y}$, it usually consists of a feature extractor $f_{\theta}$  and a classifier $\phi$.

We also assume inputs data points $(x,y)$ are drawn from a distribution $P(x\mid y,a)$ condition on attribute $a$, our diagnosis goal is to identify groups $s=(y,a)$ that has significantly more errors than images draw from the marginal distribution $P(x\mid y)$. So we define the error gap:
\begin{equation}
\begin{aligned}
    \Delta(s) &= E_{P(x| y,a)} l(h(x),y)
                -E_{P(x| y)} l(h(x),y)\\
                &>\epsilon
\end{aligned}
\label{eq:error_gap}
\end{equation}
Where $s=(y,a)$ is a group combined with category $y$ and attribute $a$, $l(,)$ is metric of model errors, such as cross-entropy or $0-1$ loss (we use), and $\epsilon$ is an independent value. The error gap $\Delta(s)$ indicates whether the attribute $a$ has a positive or negative impact on category $y$ and serves as an evaluation criterion for the attribute $a$.

For image data with group or attribute annotations, the data distributions $P(x|y,a)$ and $P(x|y)$ can be drawn from data subgroups directly. 
\subsection{Training Stage}
\textbf{Align CLIP and Vision Model by Feature Distilling.} Our goal is to equip the CLIP with buggy model knowledge utilizing feature distillation, a method of knowledge distillation that enable the student model and the teacher model have similar activation features \cite{hinton2015distilling,romero2014fitnets,heo2019comprehensive}. We use the method to align the activations (embeddings) of CLIP and pretrained buggy model, as shown in \cref{fig:framework-a}.

Next, we introduce notation for the CLIP model. Given image-text pairs $(x,x_T)$,$x\in \mathcal{X}$, $x_t\in \mathcal{X}_{T}$, image encoder $h_v:\mathcal{X}\rightarrow R^{d}$ can get image embedding $ h_v(x)$, text encoder  $h_T:\mathcal{X}_{T}\rightarrow R^d$can get text embedding $z_t = h_T(x_t)$. In addition, a trainable adaptor (multi-layer perceptron with residual connection). The adaptor $\mathcal{T}$ is trained with the following align-loss, which constrains the features $z_v = h_v (x)$ of the image encoder of CLIP aligning with the feature  $f_{\theta}(x)$ of the buggy model for the same input $x$:
\begin{equation}
\begin{aligned}
    \mathcal{L}_{align} = \frac{1}{N}\sum_{x}|f_{\theta}(x)-T\circ h_v(x)|
\end{aligned}
\label{eq:align}
\end{equation}
By sharing the classifier $\phi$ and the image encoder of CLIP, the adapter and classifier combine to form the visual \emph{proxy model} $M_v = \phi \circ T \circ h_v$, which possesses the same knowledge and bugs as the buggy model $h = \phi \circ f_{\theta}$.

\textbf{Closing the Modality Gap.}
One of the key problems in cross-modality diagnosis is to improve cross-modality transferability of the representation space of CLIP. We use mean centering, which has been shown as an effective method for closing the gap \cite{zhang2023drml,zhou2023test}.
\begin{equation}
\begin{aligned}
    z_t\leftarrow z_t  -\mu_t \\
    z_v\leftarrow z_v  -\mu_v
\end{aligned}
\label{eq:global_norm}
\end{equation}
Where the $\mu_v$ and $\mu_t$ are global mean of all image features $z_v$ and text features $z_t$ respectively. Since closing the GAP of the CLIP representation space, the proxy model can recognize the text more accurately. 

\subsection{Diagnosis Stage}
\textbf{Text-only Identify Bugs.} Leveraging the transferability of CLIP representation space, the text encoder of CLIP, the adapter and the linear classifier combine as the language proxy model $M_T =\phi\circ T\circ h_T$, it can perform text classification task like image task. The objective of the diagnosis is to identify data subgroups where the model is prone to errors. When candidate attributes $\mathcal{A}$ is obtained, we apply OpenAI CLIP’s 80 prompts to class names and attribute words to generate probe texts, e.g. \emph{a photo of \{attribute\} and \{class\}} or \emph{a photo of \{class\}}. These probe texts can calculate error-gap $\Delta(s)$ of all groups $s$ in increasing ordering and we present the Top-k error groups $\overline{S}=TopK(\Delta(s))$ for further intervention. 

\noindent\textbf{Obtain Candidate Attributes without Samples}
One of the challenges in applying the method to a realistic scenarios is obtaining task-related candidate attributes. In addition to well-known attributes as \emph{gender}, generating a candidate set often requires domain experts’ prior knowledge. Instead, relying on the rich world knowledge of Large Language Models (LLM), we propose a human-free method to acquire candidate  attributes. 

In this study, we utilize the language model Llama2-7B \cite{touvron2023llama} as corpus generator, acquiring task-relevant query (such as task-description or category names) to produce task-related visual corpus. Through keywords extraction, we obtained a set of tast-relevant attributes. More details can be found in \cref{appendix:query_llm}.

\section{Experiments}
\label{sec:experiment}
\subsection{Datasets and Implementation}
\textbf{CelebA} \cite{liu2015celeba} is a widely used face dataset for the spurious correlation problem with 40 attributes annotations. Following Sagawa et~al.'s work \yrcite{sagawa2019distributionally}, images are classified based on  hair color (Blond hair, Dark hair). Each category is further divided into two groups based on gender (male, female), with data proportions of dark hair (female): dark hair (male): blond hair (female): blond hair (male) = 44\%:41\%:14\%:1\%. \textbf{Waterbirds} \cite{sagawa2019distributionally} is dataset extensively used for spurious correlation issues in bird identification. Following Sagawa et~al.'s work \yrcite{sagawa2019distributionally}, images are classified into waterbird and landbird. There exists spurious correlation between bird categories (waterbird or landbird) and backgrounds (on land or on water). The proportions of different groups are landbird on land: landbird on water: waterbird on land: waterbird on water = 73\%:4\%:1\%:22\%.
\textbf{NICO++} \cite{zhang2023nico++} consist of six super-classes (such as mammals, birds and plants)  that we set as target labels in six different contexts (such as rock, grass, and water). Each super-class e.g. mammals corresponds various base sub-classes e.g. sheep,wolf,lion and so on. Following Yenamandra et~al. \cite{Yenamandra_2023_FACTS}, we set 95\% correlation in super-classes and contexts (bias-align: bias-conflicting = 95\%:5\%), spurious is the proportion for a given spurious context/attribute that contain the correlated majority label.

More implementation details can be found in \cref{appendix:implementation}.

\subsection{Diagnosing Spurious Correlation}
\subsubsection{Discovery biased attributes}
In this section, we identify underperforming subgroups $s=(y,a)$ by estimating the error-gap $\Delta(s)$ for various attributes $a$ per class $y$. Building on prior work, we diagnose the well-known biased attributes. Firstly, We evaluate group-wise accuracy of buggy models on test set, the results in the \cref{tab:pretrained_perform} show that the models learned gender, background and context spurious correlation. 
\begin{table}[h]
  \caption{Performance of three buggy models. In three test sets, the buggy models suffered spurious correlation and result biased accuracy. (Avg=overall accuracy,WG=worst group accuracy,BA=bias-aligned accuracy, BC=bias-conflicting accuracy}
  \label{tab:pretrained_perform}
  \vskip 0.2in
  \centering
  \begin{tabular}{@{}lcc|lc@{}}
    \toprule
     Acc(\%) & Waterbrids&CelebA&Acc(\%) &NICO++\\
    \midrule
    Avg&81.4 &95.7&BA&86.7\\
    WG&40.0&35.6&BC&75.1\\
    \bottomrule
  \end{tabular}
  \vskip -0.15in
\end{table}

\begin{figure}[h]
  \subfloat[CelebA: blond hair]{
        \label{fig:celeba-blond hair}
    \includegraphics[width=0.45\linewidth]{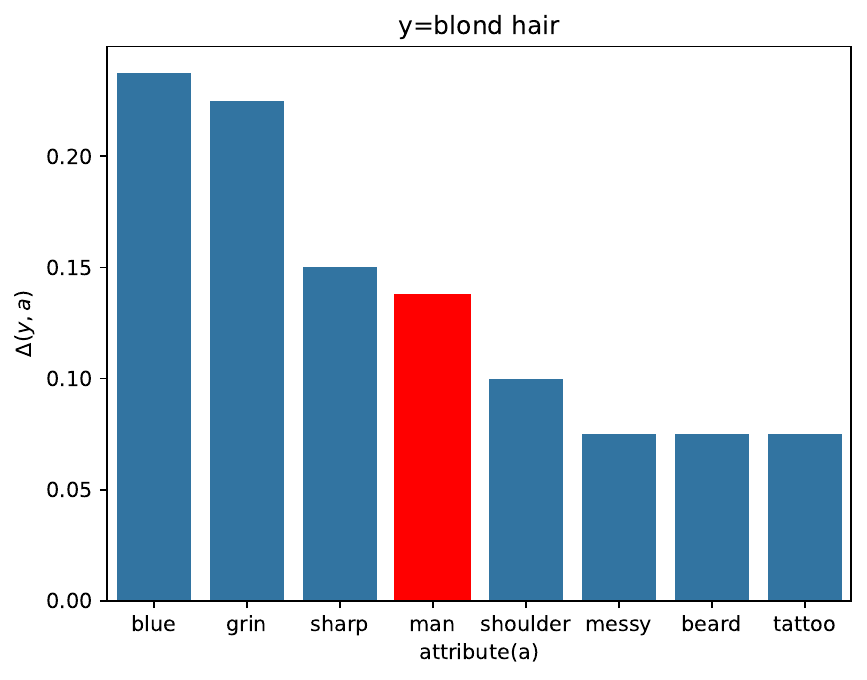}
        }\hfill
  \subfloat[Waterbirds: landbird]{
        \label{fig:waterbirds-landbird}
    \includegraphics[width=0.45\linewidth]{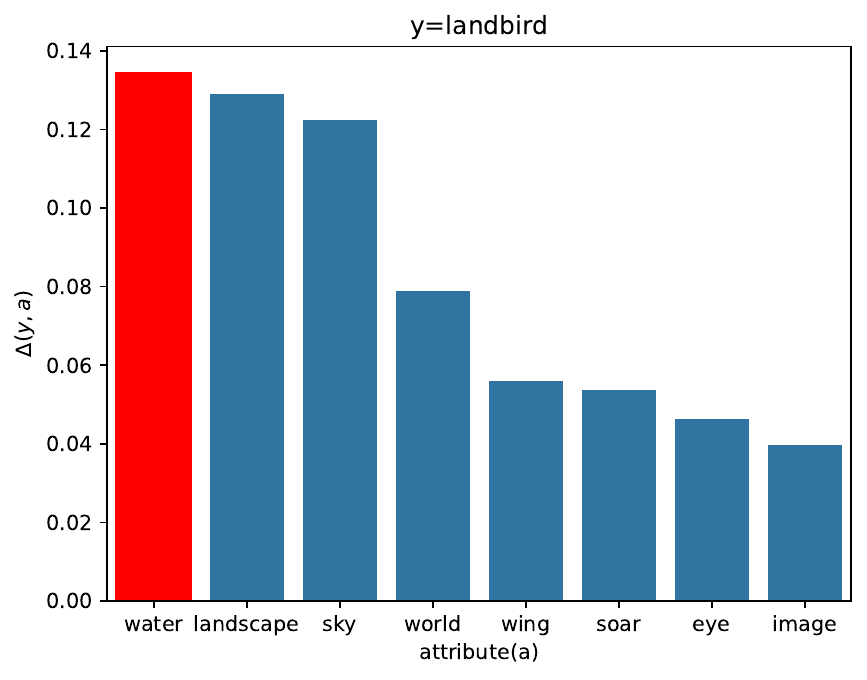}
        }\hfill
  \subfloat[Waterbirds: waterbird]{
        \label{fig:waterbirds-waterbird}
    \includegraphics[width=0.45\linewidth]{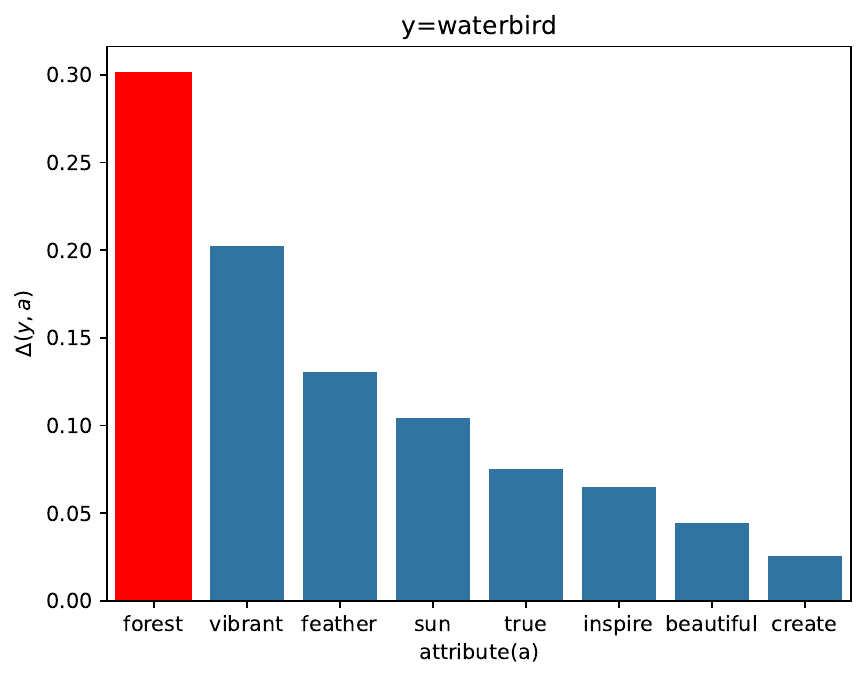}
        }\hfill
  \subfloat[NICO++: landways]{
        \label{fig:nico++-landways}
    \includegraphics[width=0.45\linewidth]{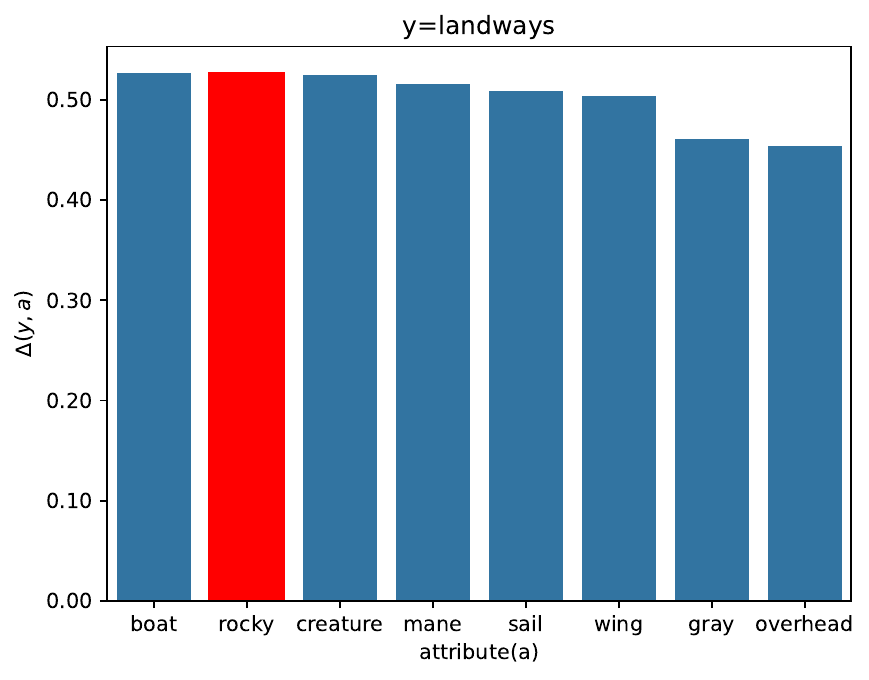}
        }    
  \caption{Discover attributes with high error gaps for each class. The attributes marked in red are the ground truth attributes annotated in three datasets. In NICO++, we only visualize attributes in the class \emph{landways}.}
  \label{fig:error-gap visulize}
\end{figure}
Furthermore, we visualize attributes with high error-gap in \cref{fig:error-gap visulize}, verify the ability of discovering well-known biased subgroups. \textbf{CelebA}: The buggy model exhibits bias towards \emph{male} within the \emph{blond hair} class. The identified biased attributes related to the ground truth attribute \emph{male} include \emph{man} and \emph{beard}. The diagnostic output also lists some other influential attributes like \emph{blue} and \emph{grin}, which represent new biases or noise. \textbf{Waterbirds}: We identified biased subgroups, such as \emph{landbird in water} and \emph{waterbird in forest} Specifically, within the \emph{landbird} category, a new attribute \emph{sky}, sharing visual semantics similar to \emph{water}, was listed. However, there were also some incomprehensible attributes generated, requiring manual filtering. \textbf{NICO++}: We identified the biased subgroup \emph{landways in rocky} and the attributes \emph{boat} and \emph{sail} are new biased attributes. Regarding the noise and incomprehensible words in the output, presumably because the adapter fails to adequately distill the visual features and the CLIP was incorrectly embedded in some words.
\subsubsection{Evaluation bias identification.}
In this section, we quantitatively evaluate the diagnostic method's capability to identify biased attributes.

\textbf{Precision@K Evaluation.} We assess LaVMD's bias labeling ability by utilizing the discovered attributes with high $\Delta(y,a)$ for the CLIP zero-shot classifier to predict the attribute annotation of each sample per-class $y$ on the test set. Following the approach of Domino \cite{Eyuboglu2022DominoDS} and FACTS \cite{Yenamandra_2023_FACTS}, we calculate Precision@K, a metric that gauges how accurately a slice discovery method can label the bias-conflicting subgroups.
\begin{equation}
\begin{aligned}
    Precision@k(A)  = \frac{1}{l}\sum_{j=1}^{l} \max_{i}P_k(a_i,\hat{a}_j)
\end{aligned}
\label{eq:precision@10}
\end{equation}
where A is the attribute labeling algorithm and $\hat{a}_j$ is a predicted attribute.
For a predicted attribute $\hat{a}_j$, we obtain a sequence of samples $x$ per class ordered by decreasing zero-shot likelihood. $P_k(a_i,\hat{a}_j)$ represents the precision of the top k samples  with ground truth attribute $a_i$. In this experiment, it measures the similarity between ground truth attribute $a_i$ and predicted attribute $\hat{a}_j$.

\textbf{Baseline}. Three slice discovery methods are under comparison. Distilling Failure Direction (FD) \cite{Jain2022DistillingMF} trains a SVM to identify model failures as directions and predict whether a sample will be misclassified by the buggy model. Domino \cite{Eyuboglu2022DominoDS} fits errors using a Gaussian-mixture-model on the validation set, extracting biased subgroups as slices. It retrieves the closest description as an interpretation for each slice. FACTS \cite{Yenamandra_2023_FACTS} improves upon Domino by first amplifying bias and then fitting a single mixture model per class instead of the entire validation set. In a similar manner, FACTS fits 36 candidate slices per class, and we designate biased candidate attributes with the 36 highest $\Delta(y,a)$ for per class.

In \cref{tab:precison-10}, our method achieves Precision@10 of 0.95, 1.0, 0.56 respectively on three test sets, matching prior work FACTS \cite{Yenamandra_2023_FACTS}. Specially, our approach does not rely on failure examples in the validation set and possesses the capability to identify attributes without real-samples. This is discussed further in the \cref{sec:Robustness2BC}.

\begin{figure}[h]
  \subfloat[Waterbirds: landbird]{
        \label{fig:waterbirds-retrieve}
        \includegraphics[width=0.45\linewidth]{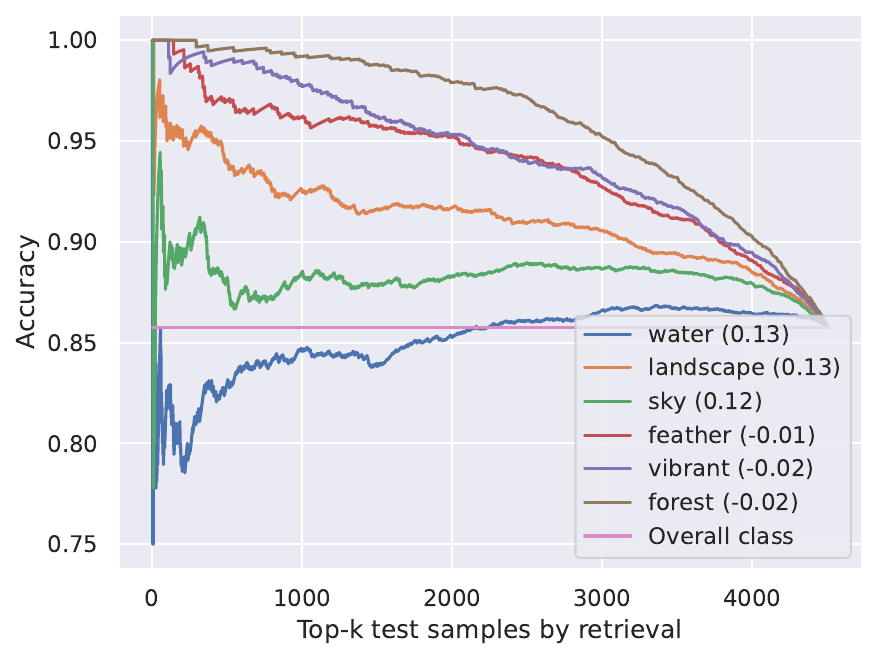}
        }\hfill
  \subfloat[NICO++: waterways]{
        \label{fig:nico++-retrieve}
        \includegraphics[width=0.45\linewidth]{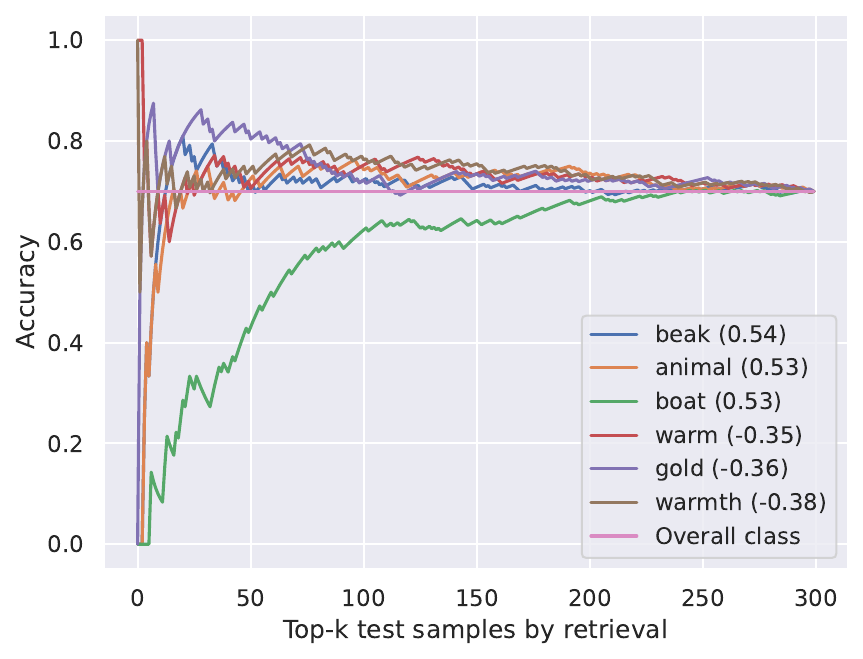}
        }
  \caption{Model accuracy  on each attribute by retrieving image samples from landbird class in Waterbirds  and waterways class in NICO++. We visualize the three attributes with 3 highest $\Delta(y,a)$ and 3 lowest $\Delta(y,a)$.}
  \vskip -0.15in
  \label{fig:retrieve samples}
\end{figure}

\begin{table}
  \centering
    \caption{Classwise precision@10 for retrievaling ground truth bias-conflicting subgroups. The results of baselines are taken from the FACTS \cite{Yenamandra_2023_FACTS}. }
    \vskip 0.2in
  \begin{tabular}{@{}llccc@{}}
    \toprule
     Method& Model&Waterbirds &CelebA&NICO++\\
    \midrule
    FD&RN50&0.9&0.7&0.19\\
    Domino&RN50&\textbf{1.0}&\underline{0.9}&0.27\\
    FACTS&RN50&\textbf{1.0}& \textbf{1.0}&\textbf{0.62}\\
    \midrule
    LaVMD&RN18&\underline{0.95}&\textbf{1.0}&\underline{0.56}\\
    \bottomrule
  \end{tabular}
    \vskip -0.15in
  \label{tab:precison-10}
\end{table}

\textbf{Evaluate on Nearest-neighbor Images.} 
Precision@10 can only match the closest predict attribute for each ground truth bias-conflicting attribute. Do these various attributes found represent real failure modes of the buggy model? Without ground-truth annotations per attribute, we can't report the accuracy on these subgroups (attributes per class). However, we can calculate image accuracy by retrieving the closest K images to an attribute within the class similar to \cite{Jain2022DistillingMF}. 

\begin{figure*}[ht]
  \centering
  \vskip 0.1in
\includegraphics[width=1.0\linewidth]{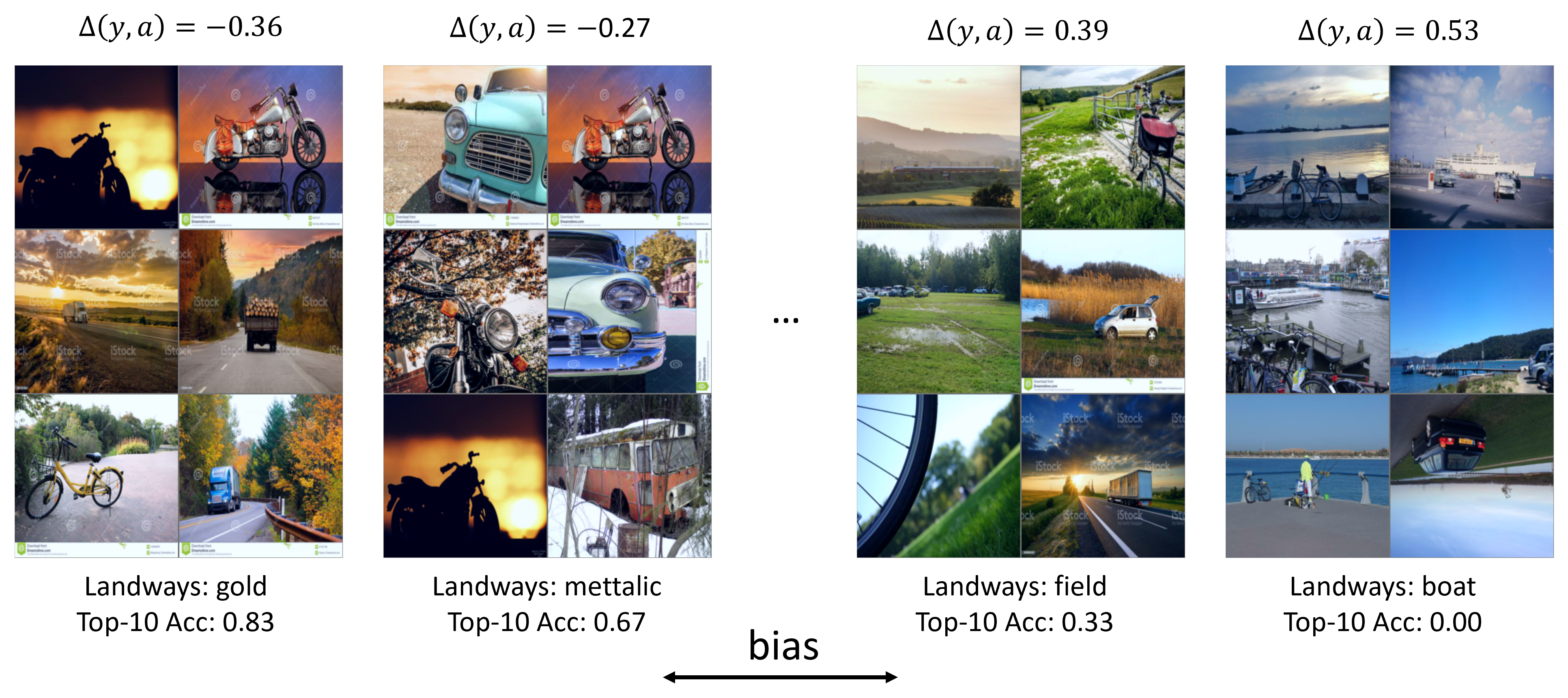}
   \caption{Visualize samples corresponding attributes. We choose several understandable attributes predicted by our method for the landways class from NICO++ and present the top-6 samples in these attribute. At the same time, we present accuracy of the top-10 samples .}
   \label{fig:slice}
   \vskip -0.15in
\end{figure*}

We order attributes by error-gap $\Delta(y,a)$ and choice several attributes with the highest/lowest error gap. A larger $\Delta(y,a)$ means higher image accuracy, vice versa. In \cref{fig:retrieve samples}, we show some attributes discovered by our method on the \emph{landbird} class of Waterbirds and the \emph{landways} class of NICO++. We can find the test images closest to these attributes with higher $\Delta(y,a)$ in CLIP embedding space correspond lower accuracy and these attributes are close to or different from ground-truth attributes. However, some biased attributes do not correspond low accuracy, presumably because some attribute words are ambiguous and the visual features are not adequately distilled by the adapter.
\cref{fig:slice} visualize retrieved top-6 samples of several attributes discovered by our method on \emph{landways} of NICO++, and calculate the top-10 samples accuracy per attribute. The \emph{landways} is spuriously correlated with \emph{autumn} context and biased with \emph{rock,grass,dim,outdoor,water}.  We observe that our method is able to discover some biased or spuriously correlated subgroups, which we name directly using attribute.

\subsection{Robustness to Bias-conflicting Samples}
\label{sec:Robustness2BC}
Our method identifies bias-conflicting attributes per class without relying on real images. That means we can discover bugs related to long-tailed data distribution, e.g. rare visual concepts. Our method can effectively leverage existing diagnostic data to generate new subgroups and infer their performance, so our method avoids additional data collection costs and is more robust to bias-conflicting samples.
We systematically decrease the proportion of bias-conflicting samples in the Waterbirds validation set (30\%, 10\%, 0\%) to assess the robustness of bias identification. We do not include a comparison with FACTS because FACTS builds upon Domino, necessitating an initial bias amplification by fitting another biased model, rather than relying on a pretrained biased model. 
\begin{figure}[h]
\vskip -0.1in
  \subfloat[Image: 0\% BC samples.]{
        \label{fig:waterbirds-umap-image}
        \includegraphics[width=0.45\linewidth]{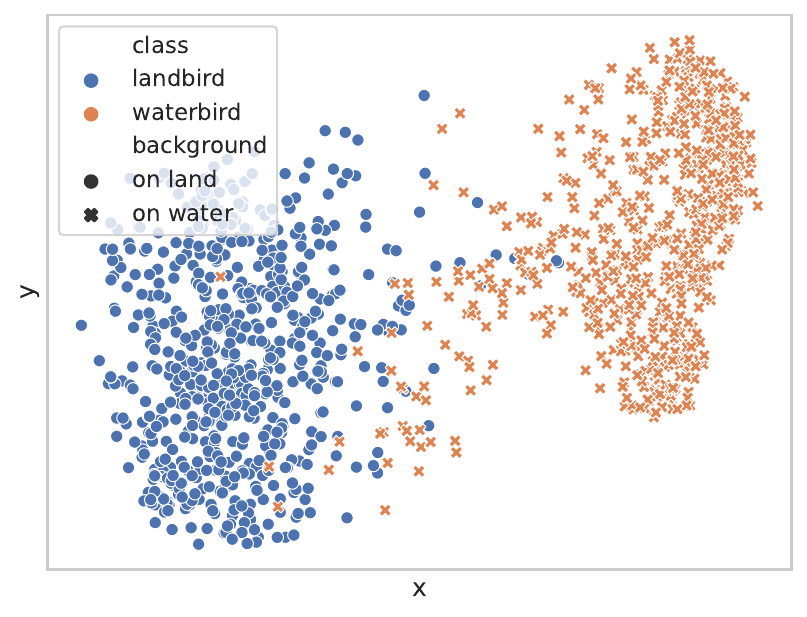}
        }\hfill
    \subfloat[Text: all samples.]{
        \label{fig:waterbirds-umap-text}
        \includegraphics[width=0.45\linewidth]{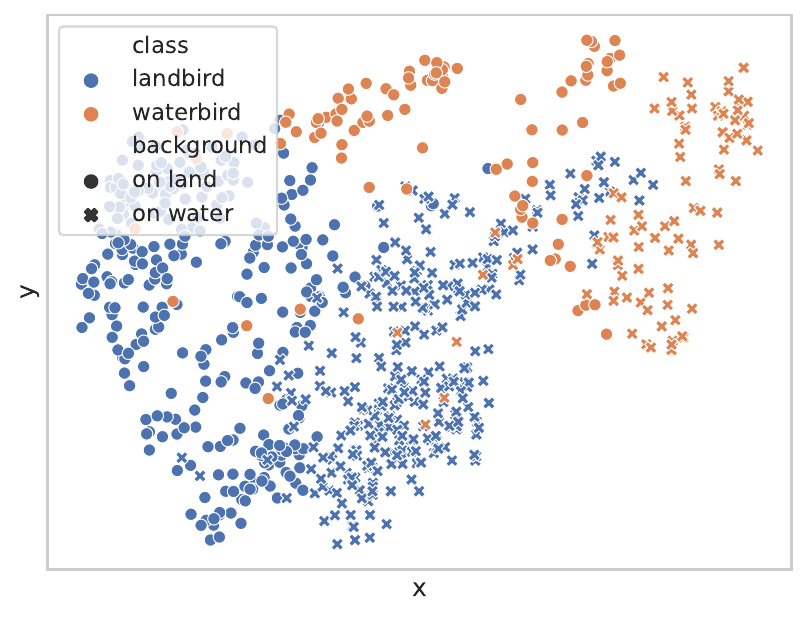}
        }\hfill
  \caption{Visualization using UMAP: (a) Image embedding of only 0\% bias-conflicting samples from the buggy model, and (b) Text embedding of descriptions for the 4 groups from CLIP+Adapter.}
  \label{fig:umap}
  \vskip -0.15in
\end{figure}
As shown in \cref{tab:precison-10-(0-30)}, Domino presents a gradual decline performance and lacks robustness. When the bias-conflicting proportion is 0\% (with only bias-aligned subgroups: \emph{landbird on land} and \emph{waterbird on water}), Domino's performance noticeably declines, because Domino fits the model predictions (probability, labels) in the validation set and relies on failed bias-conflicting examples. In fact, the original Waterbirds validation set has bias-aligned:bias-conflicting =1:1, amplifying the model error and obtains many failed examples. We emphasize that our method doesn't rely on the number of bias-conflicting visual samples. We use UMAP \cite{mcinnes2018umap} to visualize the text features of CLIP+Adapter with 0\% bias-conflicting samples.  \cref{fig:umap} presents that despite the absence of bias-conflicting samples, it still learns effective feature representations.

\begin{table}[htb]
  \caption{Classwise precision@10 for Waterbirds. On test set, the random precision@10 is 0.5. The \emph{Base} is the original data distribution.}
  \centering
  \label{tab:precison-10-(0-30)}
  \vskip 0.2in
  \begin{tabular}{@{}llcccc@{}}
    \toprule
     Method& Base&30\%&10\%&0\%\\
    \midrule    
    Domino&1.0&1.0&0.95&0.75\\
    LaVMD&0.95&0.95&0.95&0.95\\
    \bottomrule
  \end{tabular}
\vskip -0.15in
\end{table}
\subsection{Repair the Last-layer Classifier}
Inspired by DrML \cite{zhang2023drml}, our approach allows text features to be embedded into the visual representation space of the buggy model, so we can repair the last-layer classifier of buggy model using texts. We continue training the last-layer classifier of the buggy model using text data.  Two bias mitigation methods GDRO \cite{sagawa2019distributionally} and SUBG \cite{idrissi2022simple}  running on validation set are under comparison. To save memory and speed up training, we pre-extract the required image/text features and quickly repair the last layer classifier.

In \cref{tab:repair_perform}, we present the performance results of the buggy models and the repaired models. As seen, \emph{Text-only} achieves comparable or best worst-group/bias-conflicting accuracy. \emph{SUBG} and \emph{GDRO} use images and attribute annotation for model training and selection, which can be fused with text-only fine-tuning.

\begin{table*}[htbp]
  \centering
\caption{Repair buggy model by Text-only, simple finetune (\emph{ERM}) and bias mitigation methods (\emph{GRDO} and \emph{SUBG}). We continue finetune the last layer classifier of models on text or validation set, and save the model with the best worst-group (Waterbirds,CelebA) or bias-conflicting (NICO++) accuracy. Inside the parentheses is the improvements in accuracy with the buggy model.}
\vskip 0.2in
  \begin{tabular}{@{}l|cc|cc|cc@{}}
    \toprule
    \multirow{2}{*}{Acc} & 
    \multicolumn{2}{c}{Waterbirds} & \multicolumn{2}{c}{ CelebA}&\multicolumn{2}{c}{NICO++}  \\
     ~& Worst & Avg &Worst & Avg&BC&BA\\
     \midrule
     Text-only&73.7(+33.7)&88.2(+6.8)&81.6(+46.0)&85.4(-10.3)&77.7(\textbf{+2.6})&87.0(+0.3)\\
    \midrule
    ERM&81.3(\textbf{+41.3})&89.0(+7.6)&85.5(+49.9)&90.6(-5.1)&73.7(-1.4)&89.7(+3.0) \\
    GDRO&77.9(+37.9)&89.3(\textbf{+7.9})&86.7(\textbf{+51.1})&90.4(-5.3)&77.6(+2.5)&87.7(+1.0)\\
    SUBG&81.2(+41.2)&85.4(+4.0)&77.8(+42.2)&92.3(\textbf{-3.4})&69.9(-5.2)&90.0(\textbf{+3.3})\\
    \bottomrule
  \end{tabular}

  \label{tab:repair_perform}
\end{table*}

\section{Analysis}
\label{sec:ablation}
\subsection{Candidate attributes from different sources}
Consider that candidate attributes may be obtained in different ways, we compared additional two sources: task-irrelated high-frequency words from English Wikipedia\footnote{Wikipedia word frequency. \\ \href{ https://github.com/IlyaSemenov/wikipedia-word-frequency}{https://github.com/IlyaSemenov/wikipedia-word-frequency}} and  image-related  keywords from captions of images. For image-caption, we employ the pre-trained captioning model ClipCap \cite{mokady2021clipcap} for automated captioning. This simple approach can effectively generate descriptions of images, but rely on training images. The Precision@K of candidate attributes extracted from different corpora was evaluated on the NICO++ dataset (\cref{tab:precison-10-(caption)}).

The high-frequency words of the Wiki are more incomprehensible, and there is no stable diagnostic performance because there is more noise and task-irrelated words; while the words generated by LLM has a stable and better diagnostic effect, but it is still significantly worse than the high-frequency words obtained by caption. If user can get access to real images and acquire caption data, it can significantly improve diagnosis.
\begin{table}[h]
  \centering
    \caption{Classwise precision@10 for NICO++. \emph{Candi. Attri} means candidate attributions.}
    \vskip 0.2in
  \begin{tabular}{@{}c|ccc@{}}
    \toprule
     Num of 
     Condi.\
     Attri &Wiki&LLM&Image-Caption\\
    50 &0.48&0.42&\textbf{0.64}\\
    100&0.56&0.53&\textbf{0.70}\\
    200&0.45&0.56&\textbf{0.67}\\
    400&0.50&0.55&\textbf{0.60}\\
    \bottomrule
  \end{tabular}
    \vskip -0.15in
  \label{tab:precison-10-(caption)}
\end{table}
\subsection{Influence of mean centering, adapter and distill loss for cross-modal transfer}
\textbf{Mean centering and adapter.} There may be different levels of difficulty in distilling visual features on datasets for different tasks, such as face and wildlife datasets. Adapters with different hidden layers are helpful for diagnosing buggy models for different tasks. mean centering can help adapters efficiently migrate to language features. To illustrate the impact of both, we define the modality gap of the models $\mathcal{M}_A$ and $\mathcal{M}_B$ as follows by group-wise accuracy difference:
\begin{equation}
\begin{aligned}
GAP(\mathcal{M}_A,\mathcal{M}_B)={mean}_g(|Acc_g^A-Acc_g^B|)
\end{aligned}
\label{eq:group wise acc-gap}
\end{equation}

In \cref{tab:bn}, we set $\mathcal{M}_A= M_T = \phi \circ T \circ h_t$ and $\mathcal{M}_B = h = \phi \circ f_{\theta}$ and present the performance of with/without \emph{mean centering} and different hidden layer. On the CelebA, the \emph{mean centering}  obtains the lowest GAP with hidden-layer=2.
Presumably because CelebA is severe class imbalance and quite different from the wild images, \emph{mean centering} helps to adapt features in extreme scenarios. Additionally, the trained buggy model on CelebA has a specific embedding space, so the single hidden layer adapter is not enough to distill. On Waterbirds and NICO++, \emph{mean centering} obtains comparable performance with an acceptable difference. Further, we show that there is a significant difference between the feature spaces of the CLIP and the buggy model, because the linear mapping cannot align the two feature spaces.
\begin{table}[h]
  \centering
  \begin{tabular}{@{}l|cc|cc|c@{}}
    \toprule
    \multirow{2}{*}{GAP$\downarrow$} & \multicolumn{2}{c}{h-d=1}&\multicolumn{2}{c}{h-d=2}&linear  \\
    &Orig & center&Orig&center&center\\
    \midrule
    Waterbirds&\textbf{5.2}&\underline{6.4}&10.6&11.3&40.3\\
    CelebA&20.9&21.7&\underline{20.4}&\textbf{3.6}&29.3\\
    NICO++&\textbf{17.6}&\underline{19.9}&27.3&21.2&64.8\\
    \bottomrule
  \end{tabular}
  \caption{The cross-modal transferability performance of various methods for closing the modality gap. ( Orig = using the original feature; Center= mean centering, h-d = the number of adapter's hidden layers)}
  \vskip -0.15in
  \label{tab:bn}
\end{table}

\textbf{Distill loss.} In \cref{tab:distill}, we compared the effects of different knowledge distillation methods in training the adapter. We employed 3 metrics: L1, KL divergence, and L2 (Ours).
The results indicate that the L1 and L2 metrics are more effective for transferring knowledge from the buggy model to the visual branch of CLIP, with close-to-consistent performance across all subgroups.
\begin{table}[htb]
  \centering
  \begin{tabular}{@{}l|ccc@{}}
    \toprule
   GAP$\downarrow$&KL&L1`&L2\\
    \midrule
    Waterbirds&6.0&6.0&\textbf{5.4}\\
    CelebA&29.3&\textbf{3.8}&4.7\\
    NICO++&12.2&9.3&\textbf{8.2}\\
    \bottomrule
  \end{tabular}
    \caption{Comparison of different distillation methods. Three methods run with 2 hidden layer, the KL divergence sets distillation temperature=10. we compare two models $\mathcal{M}_A= M_v = \phi \circ T \circ h_v$ and $\mathcal{M}_B = h = \phi \circ f_{\theta}$  by measuring modality gap.}
  \vskip -0.15in
  \label{tab:distill}
\end{table}

\section{Conclusion}
\label{sec:conc}
This work introduces a novel method for diagnosing arbitrary vision models using text, even without relying on labeled image samples. The process involves training a feature adaptor based on the embedding space of CLIP to align it with buggy vision models. Subsequently, a Large Language Model (LLM) is employed to obtain candidate attributes, and the model's biases towards these attributes are probed for diagnosing model errors. We show that 
despite its simplicity, our method has reduced the data collection  costs of diagnosing vision models.

\nocite{langley00}

\bibliography{main}
\bibliographystyle{icml2024}

\newpage
\appendix
\onecolumn
\section{Appendix.}
\subsection{Obtain Candidate Attributes }
\label{appendix:query_llm}
We obtain the set of candidate attributes through the following steps.
\begin{itemize}
\item \textbf{Step1: Corpus were generated by querying Llama2.} We query large language models with simple questions like \cref{fig:framework-b} to generate image descriptions relevant to task. The LLM then returns a list image captions with noise, which could be cleaned through simple data preprocessing.
\item \textbf{Step2: Keyword Extraction for Candidate Attribute.} Yake \cite{campos2020yake} algorithm of keyword extraction was applied to the cleaned text to obtain a candidates attributes set.
\item \textbf{Step3: Template-based Probe Text Generation.} Descriptive text was generated by combining attributes and category names using templates, such as \texttt{a photo of {class} and {attribute}.} Subsequently, the language proxy model $M_T$ was employed to calculate the error-gap $\Delta(y,a)$.
\end{itemize}
\subsection{Implementation Details.}
\label{appendix:implementation}
In realistic scenarios, the buggy model is provided by a third-party organization and might not be acquired in the laboratory. Therefore, we fine-tune the model on the training data to learn the spurious correlation. Initially, we thoroughly fine-tune ResNet18 \cite{he2016deep_resnet} to obtain a buggy vision model on CelebA, Waterbirds, and NICO++. The training process involves the cross-entropy loss, Adam optimizer \cite{kingma2014adam} with an initial learning rate of 1e-4, and we save the model with the best overall accuracy on the validation set per epoch.

Throughout the training and diagnosis stages, we utilize CLIP-RN50 \cite{Radford2021LearningTV} and distill features from the training dataset. The structure of the adapters includes a residual-connected MLP with 1-2 hidden layers and a hidden dimension of 1024. We use the SGD optimizer with an initial learning rate of 0.1 for up to 30 epochs and implement a cosine learning rate decay until reaching 1e-3.

\subsection{Discussion}
\label{sec:discus}
\textbf{Compatibility}
In contrast to alternative methodologies relying on the generation of keywords/text associated with misidentified images or candidate attributes for error identification, our method demonstrates versatility, enabling application in a broader spectrum of scenarios. This adaptability makes it interoperable with other diagnostic solutions, and when provided access to labeled datasets or more accurate candidate attributes, it can significantly enhance diagnostic performance.

\textbf{Model Interpretability and Repair}
Our approach is a natural extension of DrML \cite{zhang2023drml} in diagnosing existing vision models. Similar to DrML \cite{zhang2023drml}, our method can evaluate concepts that significantly influence prediction results. Additionally, it can leverage text instead of images for further training the final layer classifier.

\textbf{Limitations}
While our method effectively diagnoses arbitrary vision models, its effectiveness is constrained by the capabilities of multimodal models and the availability of distillation data. For concepts not well-handled by multimodal models (e.g., digit recognition \cite{Radford2021LearningTV}), diagnostic efficacy may be limited.

\subsection{Social Impact}
This paper presents work whose goal is to advance the field of Trustworthy Machine Learning. There are many potential societal consequences of our work, none which we feel must be specifically highlighted here.

\end{document}